\pdfoutput=1

\documentclass[11pt]{article}

\usepackage[final]{acl}

\usepackage{times}
\usepackage{latexsym}

\usepackage[T1]{fontenc}

\usepackage[utf8]{inputenc}
\usepackage{wasysym}

\usepackage{microtype}

\usepackage{inconsolata}

\usepackage{graphicx}
\usepackage{multirow}
\usepackage{array}
\usepackage{booktabs}
\usepackage{tabularx} 
\usepackage{xspace}
\usepackage{gensymb}
\usepackage{makecell}
\usepackage{colortbl}
\usepackage{xcolor}
\usepackage{subcaption}
\usepackage{comment}
\usepackage[export]{adjustbox}
\newcommand{\ours}{\textsc{Mime}\xspace}
\newcommand{\adv}{\includegraphics[height=0.8em]{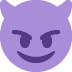}\xspace}
\newcommand{\lightbulb}{\includegraphics[height=0.8em]{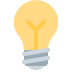}\xspace}
\newcommand{\confused}{\includegraphics[height=0.8em]{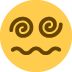}\xspace}
\newcommand{\realdataset}{\textsc{Real}\xspace}

\usepackage{pifont}
 
\usepackage{amssymb}

\definecolor{darkgreen}{RGB}{0, 119, 187}
\definecolor{lightgreen}{RGB}{102, 178, 224}
\definecolor{darkred}{RGB}{204, 102, 0}  
\definecolor{lightred}{RGB}{255, 178, 102}

\title{Can Vision Language Models Understand Mimed Actions?}

\author{
    Hyundong Cho$^1$\quad\;
    Spencer Lin$^2$\quad\;
    Tejas Srinivasan$^3$\quad\; 
    Michael Saxon$^4$ \\
    \textbf{
    Deuksin Kwon$^2$\quad\; 
    Natali T. Chavez$^5$\quad\;
    Jonathan May$^1$}~\\
    Information Sciences Institute$^1$, Institute for Creative Technologies$^2$, Department of Computer Science$^3$ \\ 
    University of Southern California \\
    University of California, Santa Barbara$^4$ \xspace
    Aristotle University of Thessaloniki$^5$
    \\
     {\small 
    \texttt{hd.justincho@gmail.com}
    }\\
}

\begin{document}
\maketitle
\begin{abstract}

Nonverbal communication (NVC) plays an integral role in human language, but studying NVC in general is challenging because of its broad scope and high variance in interpretation among individuals and cultures.
However, mime---the theatrical technique of suggesting intent using only gesture, expression, and movement---is a subset of NVC that consists of explicit and embodied actions with much lower human interpretation variance. 
We argue that a solid understanding of mimed actions is a crucial prerequisite for vision-language models capable of interpreting and commanding more subtle aspects of NVC. 
Hence, we propose Mime Identification Multimodal Evaluation (\ours), a novel video-based question answering benchmark comprising of 86 mimed actions. 
Constructed with motion capture data, \ours consists of variations of each action with perturbations applied to the character, background, and viewpoint for evaluating recognition robustness.  
We find that both open-weight and API-based vision-language models perform significantly worse than humans on \ours, motivating the need for increased research for instilling more robust understanding of human gestures.%

\end{abstract}

\section{Introduction}
\label{sec:introduction}

\begin{figure}[t]
    \centering
    \includegraphics[width=\linewidth]{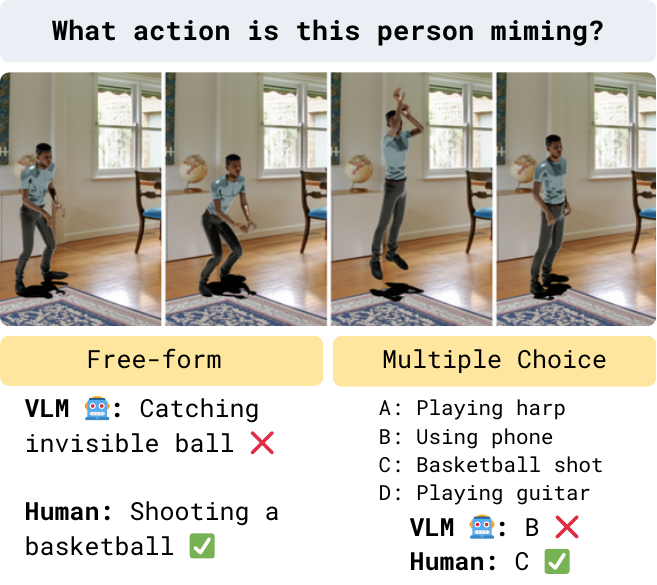}
    \caption{
    Simplified illustration of a sample in \ours shown with a few frames from a video of a 3D male character miming a basketball shot in a living room. 
    Humans achieve almost perfect accuracy on identifying mimed actions regardless of evaluation format, adversarial perturbations, and the absence of salient context (e.g., basketball, court, basketball outfit), while VLMs struggle without salient context. 
    }
    \vspace{-1em}
    \label{fig:intro_figure}
\end{figure}

\begin{figure*}[t]
    \centering
    \includegraphics[width=\linewidth]{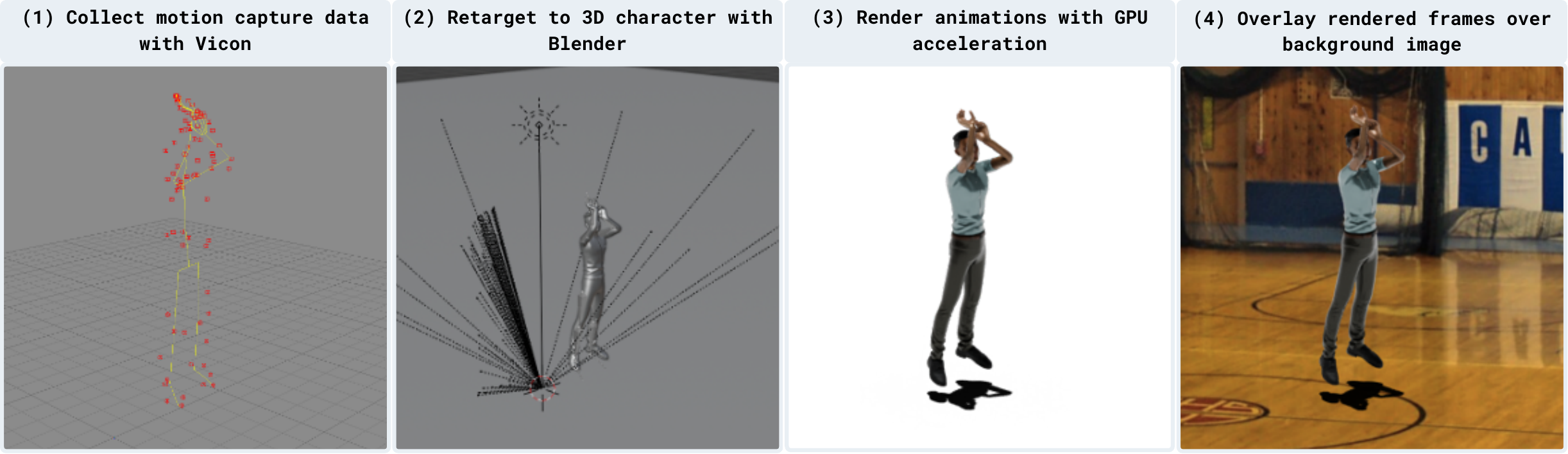}
    \caption{An overview of the pipeline for constructing \ours. \texttt{(1)} We first collect motion capture data of a mimed action on a Vicon stage. \texttt{(2)} Then, a 3D character is retargeted to our motion capture data in Blender, a computer graphics software. \texttt{(3)} Next, we render frames of the animation with a transparent background. \texttt{(4)} With frames rendered with transparent backgrounds, we can easily overlay them over images of our choice.
    }
    \vspace{-1em}
    \label{fig:data-collection-overview}
\end{figure*}

Nonverbal communication (NVC) --- the use of nonverbal cues such as gestures, facial expressions, and body language to convey messages --- is an instrumental part of human language~\cite{mehrabian1972nonverbal, poyatos1983language, stickley2011soler}.
NVC not only serves as a crucial substitute to communication when verbal modes are limited~\cite{friedman1979nonverbal, mast2007importance, park-etal-2022-pictalky, shafique2023nonverbal, karmakar-sinha-2024-aiding}, but also makes interaction engaging and natural~\cite{kendon1967some, duncan1969nonverbal, ha-etal-2012-combining, xu-etal-2022-gestures}, and may even betray true intent that contradicts what is verbally expressed~\cite{mehrabian1972nonverbal, mcneill1992hand, eaves2015successful}.
Therefore, AI systems need to establish a thorough understanding of NVC for them to become more accessible and effective assistants to humans~\cite{argyle1979person, troshani2021we}. 

Unfortunately, this is an overwhelming undertaking considering the broad scope of NVC~\cite{mehrabian1972nonverbal, eaves2015successful}, variability in how individuals interpret and exhibit nonverbal cues~\cite{kita2009cross, matsumoto2013cultural}, and the limited capabilities of current vision-language models (VLMs)~\cite{radford2021learning, xu-etal-2021-videoclip, chen2024internvl, abdin2024phi3technicalreporthighly, Qwen-VL, geminiteam2024gemini15unlockingmultimodal, tang2025video}. 
Despite impressive achievements of VLMs on action recognition benchmarks~\cite{kong2022human, wang2023actionclip, qu2024llms}, we find that they cannot even reliably identify a subset of NVC that human adults without apraxia\footnote{A neurological disorder that disrupts the ability to plan and execute purposeful movements, despite having the physical ability to do so.} comprehend with ease~\cite{o1995using}: mime, the theatrical technique of suggesting intent using only gesture, expression, and movement. 
Compared to other general gestures, many mimed actions are consistently identified among humans, in part due to their direct ties to physical movement and surfaces~\cite{o1995using, alexanderson2017mimebot, van2017production, little2021physically}. 
Therefore, we propose studying whether VLMs can reliably recognize mimed actions as a foundational prerequisite towards the sophisticated comprehension of the full spectrum of NVC. 

To this end, we address the following research questions: (\textit{i}) \textit{Can VLMs reliably recognize mimed actions?} and (\textit{ii}) \textit{If not, can we improve a VLM's performance on identifying mimed actions?} 
For the first research question, we construct \textbf{M}ime \textbf{I}dentification  \textbf{M}ultimodal \textbf{E}valuation (\ours),\footnote{Data and code for \ours is available \url{https://justin-cho.com/mime}.
}  a novel video-based question answering benchmark comprising of 86 mimed actions. 
We create \ours using motion capture data and computer graphics software, which enables us to create variations of each action with perturbations applied to the character, background, and viewpoint for evaluating recognition robustness (see \autoref{fig:intro_figure} for a sample of \ours and corresponding human and VLM predictions). 
On \ours, humans achieve almost 100\% accuracy, regardless of adversarial perturbations and evaluation format.
However, VLMs, open-weight models and API-based black-box models alike, only achieve at most 52.3\% accuracy in a multiple choice format where contextual information is provided by the answer choices and at most 19.8\% with a free-from short answers format. 
Accuracy is even lower for videos with adversarial perturbations, for which all evaluated models achieve less than 10\%.  
On the other hand, their performance is significantly boosted when provided a background that is contextually relevant (e.g., basketball court for mime of basketball shot). 

To answer the second research question, we conduct a preliminary exploration into whether existing methods can bridge this shortcoming. 
Specifically, we experiment with Chain of Thought~\cite{wei2022chain}, few-shot in-context learning, and fine-tuning with a subset of \ours. 
We find that the only method that consistently improves model performance over zero-shot is few-shot in-context learning for API-based black-box models, but their results remain significantly worse than human performance.
A manual inspection into the descriptions of the mimed actions generated by using Chain of Thought with Gemini 1.5 Flash reveal that the majority of failure cases is due to incorrect observations of the demonstrated gestures (80\%) and a smaller portion is from incorrectly interpreting correctly generated descriptions (15\%). 
In conclusion, our findings with \ours motivate research that instills a more robust understanding of human gestures in VLMs for establishing an essential foundation for NVC comprehension.

\begin{figure*}[th!]
    \centering
    \includegraphics[width=\linewidth]{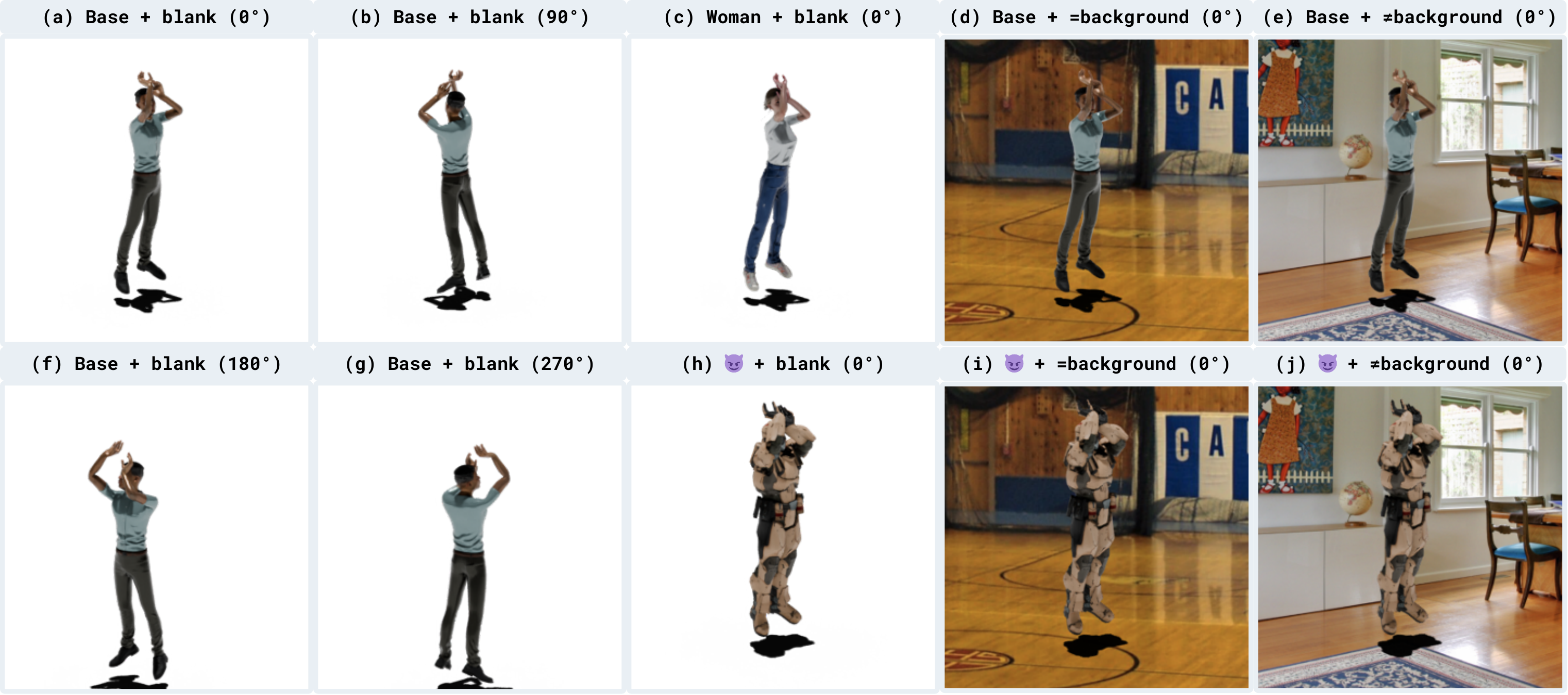}
    \caption{Overview of variations of each action in \ours. Our setup of using motion capture and computer graphics software allows us to flexibly permute different configurations for each action to ablate the robustness of a VLM's understanding of mimed actions. \texttt{(a,b,f,g)} are examples of the same animation but with changes to the camera angle where different body parts become occluded depending on the angle. \texttt{(c)} and \texttt{(h)}  only change the character from \texttt{(a)}. \texttt{(c)} is a female human character while \texttt{(h)} is an adversarial character \adv in a sci-fi spacesuit. \texttt{(d)} and \texttt{(i)} are variants of \texttt{(a)} and \texttt{(h)} respectively with aligned backgrounds (\texttt{$=$background}, e.g., basketball court for basketball-related action) while \texttt{(e)} and \texttt{(j)} have adversarial backgrounds (\texttt{$\neq$background}, e.g., living room).
    }
    \vspace{-0.5em}
    \label{fig:mime-dataset-overview}
\end{figure*}

\section{\ours}
\label{sec:mime_dataset}

In this section, we describe the data collection pipeline for \ours.
An overview is shown in \autoref{fig:data-collection-overview}.
\ours is a video-based question answering benchmark that comprises of animations of 86 mimed actions, each with ten variants that are shown in \autoref{fig:mime-dataset-overview}, resulting in a total of 860 evaluation samples.
The videos are rendered with 3D graphics software by combining digital assets with motion capture data of actors miming various actions. 
This setup is advantageous to alternative methods\footnote{We discuss challenges with alternative methods, such as using live action footage and video generation models in Appendix \ref{appdx:alternative_generation}.} for conducting a systematic study of recognition robustness with regards to various components that comprise an action as each action can be post-processed and remixed with different backgrounds, characters, and camera angles. 

\subsection{Collecting Motion Capture Data}

First, we brainstorm 75 mimed action candidates for which salient context is missing. 
For example, playing a violin is a valid candidate because it is acted out without a violin and swimming is also valid because it is acted out without being in water, and both mimed actions are understood by human subjects.
On the other hand, we exclude gestures such as hand-waving or thumbs-up as no salient context is missing in their enactment. 

Next, we have two actors (one male nonprofessional actor and one female professional actor) act out these action candidates with three takes each. 
Each take introduces some variance of the same acts if there are multiple ways to perform them (e.g., swimming can be done with front stroke, back stroke, etc. and pushing can be done with various intensity) and if they are clearly distinct, multiple takes of the same action are kept. 
For more complex actions such as shot putting, the actors reference YouTube videos of professional athletes. 

Only the motion capture data for which at least two out of three authors assign the same label to the final rendered output without seeing the action name are included in \ours.
This process results in 47 action types and 86 mimed action samples. 
Additional technical details of our motion capture process is described in Appendix \ref{appdx:motion_capture_details}.

\subsection{Creating Blender Files}

Motion capture data is imported into Blender and combined with digital assets to render frames with a transparent background so that they can be easily overlaid over our background of choice later without redundant rendering.

To efficiently combine various characters with a large number of motion capture data together, we write a Python-based macro that automates the process of creating blender files to be rendered. 
The result of the macro is shown in \texttt{(2)} of \autoref{fig:data-collection-overview}.
The detailed steps that our script automates are elaborated in Appendix \ref{appdx:blender-processing}.

\subsection{Rendering with Variations}

\paragraph{Characters}
We use free 3D characters from Mixamo.\footnote{\url{https://www.mixamo.com}} 
For the base setting of \ours, we use a male human character with casual clothes.
To evaluate for mime recognition robustness with regards to the character, 
we also render with an adversarial character that is wearing a sci-fi spacesuit (shown in \texttt{(h,i,j)} in \autoref{fig:mime-dataset-overview}. 
While we may choose even more adversarial characters that look less human to create a more challenging variant, we find that not all motion capture data is compatible for characters with largely diverging body proportions as the mimed action can become unrecognizable due to different body parts overlapping one another. 

To test for a VLM's robustness to the character's gender, we also render with a female human character with casual clothes. 
The female character that we use is illustrated in \texttt{(c)} in \autoref{fig:mime-dataset-overview}.

\paragraph{Backgrounds} 

We use images from Creative Commons licensed images from Wikimedia\footnote{\url{https://commons.wikimedia.org/}} as aligned and misaligned backgrounds (e.g., \texttt{(d,i)} and \texttt{(e,j)} in \autoref{fig:mime-dataset-overview}, respectively).  
We do our best to find images for which the background provides a large open space in the middle so that the full action sequence does not look awkward and the character does not appear disproportionately large or small.\footnote{While most images fulfill this criteria, there are a few for which it was not feasible to scale or crop properly so that the character ends up disproportionately large, such as the example shown in \autoref{fig:human_evaluation_interface} in Appendix \ref{appdx:human_evaluation_details}. 
However, we find this not to be an issue for humans to correctly identify the mimed action, and therefore consider reasonable evaluation samples and keep them in \ours.  }

\paragraph{Angle}

To test robustness to viewpoints of the observed mimed action, we also render videos with various angles by rotating the camera with the character at the center. 
We select angles of 90\degree, 180\degree, and 270\degree\xspace rotations applied to the base setting. 
These are shown in \texttt{(b,f,g)} in \autoref{fig:mime-dataset-overview}.

\section{Experimental Setup}
\label{sec:experiments}

\subsection{Evaluation Format} 
\label{ssec:evaluation}

Prior work examine mime recognition under two different question answering conditions, the choice condition and naming condition~\cite{osiurak2012make, van2017production}, as the results between the two can differ significantly. 
The choice condition provides answer choices, which in effect supplies contextual information, while the latter requires answering directly without any choices and is therefore more challenging and leads to lower agreement. 
Therefore, we construct \ours so that it evaluates VLMs with both of these conditions. 
We elaborate on the setup for each condition in the following. 

\paragraph{Choice condition: multiple choice (MC)} This is the best setting for computing accuracy as it can be done with exact match, but performance is dependent on how confusing the distractors are. 
Our multiple choice setup has four options to choose from and the distractors are selected by randomly sampling from other action labels that are included in \ours after removing the top 10 that have highest cosine similarities when compared with sentence embeddings~\cite{reimers-gurevych-2019-sentence}.\footnote{\texttt{sentence-transformers/all-MiniLM-L6-v2}} 
While this may make the multiple choice setup easier, it simplifies evaluation by preventing instances where there are multiple valid answers.

\paragraph{Naming condition: Free-form short answers (FF)}
In order to test model performance when it is not provided any context from the multiple choice options, we also assess their performance with a free-form short answer format. 
To assess the reference-based accuracy of our freeform answers, we adopt a single sentence-embedding cosine-similarity-based metric, effectively a relaxation of BertScore \cite{zhang2019bertscore}, which is popular in VLM question answering-based evaluation of text-image similarity \cite{hu2023tifa,saxon2024evaluates}. 
We use a sentence transformers model, the same one used for selecting distractors in the multiple choice format, to produce sentence-level embeddings of the generated free-form answers and gold labels, and use a heuristically-selected cosine similarity threshold of 0.5 to mark an answer as correct. 
While we find these to return a few false positives (e.g., baseball swing given credit for baseball pitch) and false negatives (e.g., pulling not given credit for dragging), we find these to be a small subset that does not significantly shift the overall performance of a model.

\subsection{Models}
\label{ssec:models}

We evaluate a comprehensive set of open- and closed-source VLMs with \ours to get a general understanding of whether VLMs can identify mimed activities. 

For open-source models, we evaluate on 
(\textit{i}) Qwen 2.5 VL Instruction ~\cite{Qwen2.5-VL}, both 3B and 7B versions, (\textit{ii}) InternVL 2.5 8B Instruct ~\cite{chen2024internvl}, (\textit{iii}) Phi 3.5 VL Instruction, which is a 4.2B model released by Microsoft ~\cite{abdin2024phi3technicalreporthighly}. 
For closed-soure models, we evaluate on (\textit{iv}) Gemini 1.5 Flash from Google ~\cite{geminiteam2024gemini15unlockingmultimodal} and (\textit{v}) GPT-4o Mini from OpenAI.\footnote{\texttt{gpt-4o-mini-2024-07-18}, \url{https://platform.openai.com/docs/models/gpt-4o-mini}} 
For our first set of results, we use a zero-shot setting where the models are asked to directly predict the answer based on the video without any examples or reasoning steps. 
Our zero-shot prompt for multiple choice and free-form formats are shown in Appendix \ref{appdx:prompt_details}. 

\subsection{Human Evaluation}
\label{ssec:human_eval}

We measure human performance on \ours to ensure that \ours is a tractable benchmark that humans perform well on and also confirm prior research that mimed action has low interpretation variability among humans~\cite{o1995using, alexanderson2017mimebot, van2017production, little2021physically}.   
We recruit 60 internal participants from the University of Southern California's Viterbi School of Engineering.
They cover a wide demographic with eight unique nationalities, ages ranging from early 20s to mid 40s, and a 6:4 ratio of men to women.
Although most are located in the same city, we believe their diverse international backgrounds provide a reasonable representation of general human performance.
Each sample across all variations in \ours is completed by three participants.  
We share further detail on our human evaluation setup in Appendix \ref{appdx:human_evaluation_details}.

\begin{figure}[t]
    \centering
    \includegraphics[width=0.8\linewidth]{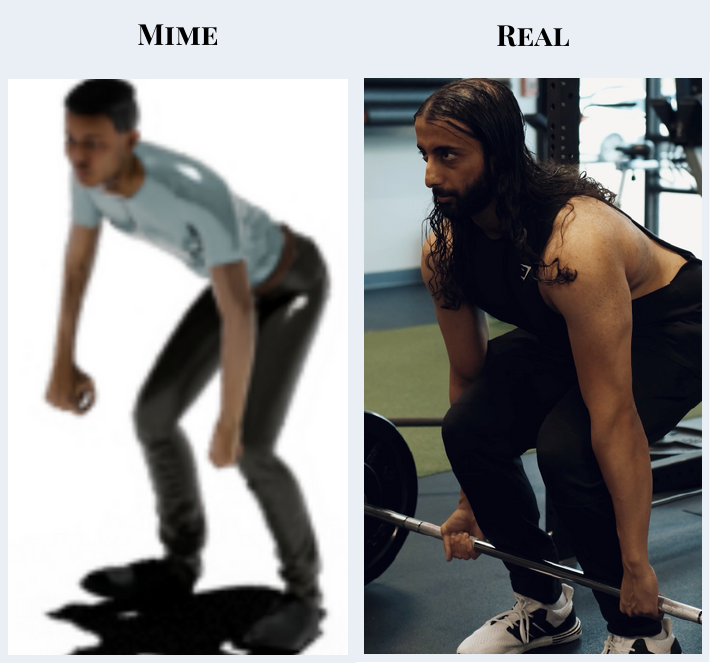}
    \vspace{-0.5em}
    \caption{A frame from videos of deadlifting from \ours (left) and \realdataset (right). In \ours, salient context is missing (e.g., barbell and gym clothing).}
    \vspace{-1em}
    \label{fig:mime_vs_real}
\end{figure}

\subsection{\realdataset}
\label{ssec:corresponding_real_data}

We ground the performance on \ours by measuring the performance on recognizing actions from live action footage (i.e., video created through traditional filmmaking techniques, capturing real actors, props, sets, and locations.) of the same set of actions in \ours. 
We collect a set of royalty- and copyright-free videos of such footage sourced from Pexels\footnote{\url{https://www.pexels.com/}} and call it \realdataset. 
An example of a video from \realdataset and its corresponding sample in \ours is shown side by side in \autoref{fig:mime_vs_real}. 
\realdataset functions as a control dataset that estimates a VLMs understanding of the actions that are mimed in \ours when all reasonable salient context is present. 
Therefore, the gap between performance on \realdataset and \ours serves as a proxy in the lack of generalizability in the understanding of the action to the understanding of its mimed counterpart. 
Note that while \ours contains 86 total mimed actions with multiple variations of the same activity, we only find one for each in \realdataset, and therefore \realdataset consists of 47 videos.

\begin{figure*}
    \centering
    \includegraphics[width=\linewidth]{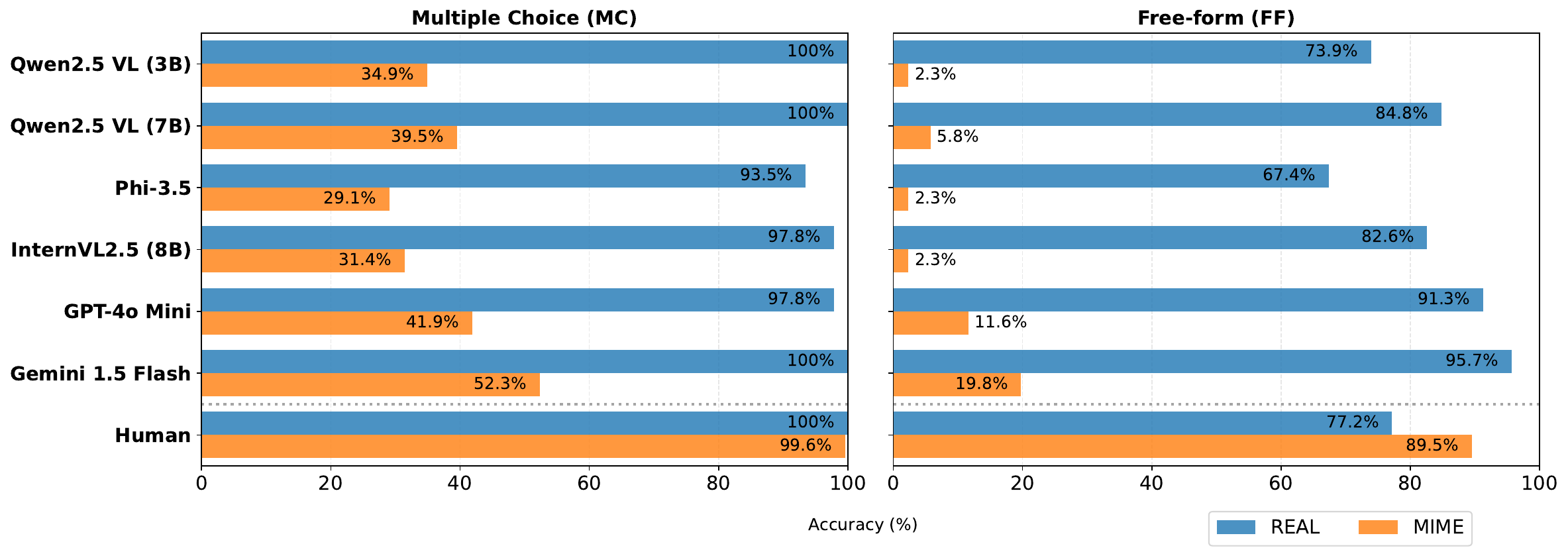}
    \vspace{-2em}
    \caption{Performance comparison on the base setting of \ours and on the \realdataset dataset. Humans show equally strong performance on both \ours and \realdataset. VLMs struggle with \ours while achieving comparative performance on \realdataset, which suggests they lack a robust understanding of human actions.
    }
    \label{fig:mime_vs_real_results}
\end{figure*}

\begin{table*}[h]
    \centering
    \adjustbox{max width=\textwidth}{
    \begin{tabular}{lcc|cccccccccc}
        \toprule
        \multirow{2}{*}{\textbf{Model}}  & \multicolumn{2}{c}{\textbf{\texttt{Base $+$ blank}}} & \multicolumn{2}{c}{\textbf{\texttt{Base $+$ $=$back.}}} & \multicolumn{2}{c}{\textbf{\texttt{Base $+$ $\neq$back.}}} & \multicolumn{2}{c}{\textbf{\texttt{\adv $+$ blank}}} & \multicolumn{2}{c}{\textbf{\texttt{\adv $+$ $=$back.}}} & \multicolumn{2}{c}{\textbf{\texttt{\adv $+$ $\neq$back.}}} \\
        \cmidrule(lr){2-3} \cmidrule(lr){4-5} \cmidrule(lr){6-7} \cmidrule(lr){8-9} \cmidrule(lr){10-11} \cmidrule(lr){12-13}
        & \textbf{MC} & \textbf{FF} & \textbf{MC} & \textbf{FF} & \textbf{MC} & \textbf{FF} & \textbf{MC} & \textbf{FF} & \textbf{MC} & \textbf{FF} & \textbf{MC} & \textbf{FF} \\
        \midrule
Qwen 2.5 VL (3B)   & 34.9 & 2.3 & 61.6 & 30.2 & 27.9 & 0.0 & 30.2 & 1.2 & 60.5 & 29.1 & 24.4 & 0.0 \\
Qwen 2.5 VL (7B)  & 39.5&5.8&\underline{68.6}&38.4&32.6&1.2&34.9&0.0&64.0&30.2&30.2&0.0  \\
Phi 3.5  & 29.1 & 2.3 & \textbf{73.3} & 27.9 & 31.4 & \underline{8.1} & \textbf{44.2} & 0.0 &  \underline{72.1} & 27.9 & \textbf{36.1} & \textbf{5.8} \\
InternVL2.5 8B  & 31.4 & 2.3 & 57.0 & 26.7 & 22.1 & 2.3 & 25.6 & 2.3 & 59.3 & 20.9 & 30.2 & 2.3 \\ \arrayrulecolor[gray]{0.8} \midrule
GPT-4o Mini  & \underline{41.9} & \underline{11.6} & 66.3 & \underline{39.5} & \textbf{37.2} & 3.5 & 33.7 & \textbf{8.1} & 67.4 &  \underline{33.7} & \textbf{36.1} & 2.3 \\
Gemini 1.5 Flash & \textbf{52.3} & \textbf{19.8} & \underline{68.6} & \textbf{51.2} & \textbf{37.2} & \textbf{12.8} & \textbf{44.2} & \textbf{8.1} & \textbf{75.6} & \textbf{46.5} & \textbf{36.1} &  \underline{3.5} \\ \arrayrulecolor{black} \midrule 
Human & 99.6 & 89.5 & 98.5 & 89.2 & 99.2 & 93.4 & 98.5 & 93.8 & 99.2 & 94.1 & 99.2 & 95.0 \\
        \bottomrule
    \end{tabular}
    }
    \caption{Evaluation results on \ours for various perturbations. 
    We use the same notations as \autoref{fig:mime-dataset-overview}, with \texttt{back.} used as a shorthand for \texttt{background}.
    Refer to \autoref{fig:mime-dataset-overview} to view samples of each variation.
    Humans are robust to all variations, but VLMs drop performance for adversarial perturbations and get a significant boost when exposed to signals from the background that are aligned with the action. 
    }
    \vspace{-1em}
    \label{tab:main-perturbations}
\end{table*}

\section{Results}
\label{sec:results}

\subsection{\ours vs \realdataset}
\label{ssec:base_results}

\paragraph{Humans understand actions and their mimed counterparts equally well, while VLMs struggle significantly for the latter.}
First, we share our results with the models mentioned in Section \ref{ssec:models} on the base setting of \ours (\texttt{(a)} in \autoref{fig:mime-dataset-overview}) and \realdataset in 
\autoref{fig:mime_vs_real_results}. 

Results on \realdataset clearly indicate that all VLMs are able to identify actions when all of the salient context is present (e.g., doing a deadlift in a gym with a barbell while wearing gym attire), achieving almost perfect scores for the MC while showing only a minor drop for the FF.
This is on par with human performance. 

However, on \ours, VLM performance drops sharply, while human performance remains consistent, with only a 0.4\% drop in MC while there is a boost for FF by 12.3\%. 
Upon manual inspection, we find that this is not because human performance is worse with live action footage, but rather because humans are more descriptive in their responses for FF for \realdataset and this produces more false negatives. 
Gemini 1.5 Flash shows the strongest performance, but even its accuracy is slightly over 50\% in MC and less than 20\% in FF. 

\subsection{Character and Background Variations}
\label{ssec:character_background_perturbations}

\paragraph{Humans demonstrate similar performance across all variations, while VLMs benefit from contextual hints and suffer from adversarial perturbations.}
The main advantage of \ours is the flexibility to swap out components of the animations in order to conduct ablation studies that shed light on the nature of the VLMs shortcomings. 

We apply the perturbations shown in \autoref{fig:mime-dataset-overview} to test how performance is affected when the character and backgrounds are changed. 
Results from these perturbations with zero-shot are shown in \autoref{tab:main-perturbations}. 

The most noticeable result from this table is that the aligned background significantly boosts performance, even when the character is adversarial. 
With the direct opposite effect, changing the background to an adversarial one seriously harms performance for most models, but interestingly less so for the open-weight models. 
Interestingly, humans are extremely robust to all of the given perturbations, maintaining almost perfect scores on all MC settings while scoring at least 89.5\% in the FF settings. 
These results indicate that while humans are able to ignore irrelevant information and focus on the actions themselves, VLMs rely on other hints about the action present in the scene.  
These results are in line with the resuls on \realdataset. 

\subsection{Angle and Gender Variations}
\label{ssec:angle_perturbations}

\paragraph{VLMs demonstrate higher variance across angle and gender variations than humans.}
Next, we share results with various angle perturbations to observe whether VLMs are viewpoint-agnostic for identifying mime. 
We see that humans clearly are consistent in this setting as well, as shown by the small variance in scores in the last row of \autoref{tab:angle_variation}. 
For the most part, \ours is challenging such that performance remains low regardless of the angle and there is no clearly preferred angle shared by VLMs. 
However, for MC, the variance in accuracy is much larger for VLMs than humans, another indication of a lack of robustness in VLMs in comparison to humans.

\begin{table}[t]
    \centering
    \begin{adjustbox}{width=\linewidth}
    \begin{tabular}{lrrrrrrr}
        \toprule
        \multirow{2}{*}{\textbf{Model}}& \multirow{2}{*}{\textbf{Eval}}& \multicolumn{4}{c}{\textbf{Rotation Angle}} & \multirow{2}{*}{\textbf{Avg.}} & \multirow{2}{*}{\textbf{Std.} $\downarrow$} \\
        \cmidrule(lr){3-6}
         &  & \textbf{0$\degree$} & \textbf{90$\degree$} & \textbf{180$\degree$} & \textbf{270$\degree$} & & \\
        \midrule
        \multirow{2}{*}{\makecell[l]{Qwen 2.5 VL \\ (3B)}} & MC & 34.9 & 34.9 & 32.6 & 32.6 & 33.7 & 1.2 \\
               & FF & 2.3 & 1.2 & 0.0 & 1.2 & 1.2 & 0.8 \\
        \multirow{2}{*}{\makecell[l]{Qwen 2.5 VL \\ (7B)}} & MC & 39.5 & 39.5 & 50.0 & 43.0 & 43.0 & 4.3 \\
               & FF & 5.8 & 7.0 & 3.5 & 8.1 & 6.1 & 1.7 \\
        \multirow{2}{*}{\makecell[l]{Phi 3.5}} & MC & 29.1 & 31.4 & 33.7 & 33.7 & 32.0 & 1.9 \\
        & FF & 2.3 & 5.8 & 3.5 & 3.5 & 3.8 & 1.3 \\
        \multirow{2}{*}{{\makecell[l]{InternVL2.5 \\ (8B)}}} & MC & 31.4 & 36.0 & 33.7 & 37.2 & 34.6 & 2.2 \\
        & FF & 2.3 & 7.0 & 7.0 & 4.7 & 5.2 & 1.9 \\
        \arrayrulecolor[gray]{0.8} \midrule
        \multirow{2}{*}{GPT-4o Mini} & MC & 41.9 & 47.7 & 43.0 & 47.7 & 45.1 & 2.6 \\
        & FF & 11.6 & 15.1 & 13.9 & 13.9 & 13.7 & 1.3 \\
        \multirow{2}{*}{{\makecell[l]{Gemini 1.5 \\ Flash}}} & MC & 52.3 & 47.7 & 52.3 & 53.5 & 51.5 & 2.2 \\
        & FF & 19.8 & 18.6 & 17.4 & 23.3 & 19.8 & 2.2 \\
        \arrayrulecolor{black}
        \midrule
        \multirow{2}{*}{Human} & MC & 99.6 & 98.8 & 98.8 & 98.7 & 99.0 & 0.4 \\
        & FF & 89.5 & 95.0 & 90.7 & 85.1 & 90.1 & 3.5 \\
        \bottomrule
    \end{tabular}
    \end{adjustbox}
    \caption{Performance on \ours for varying angles. For MC, relative to human performance, model performance varies largely depending on the viewpoint angle.
    }
    \vspace{-0.5em}
    \label{tab:angle_variation}
\end{table}

Lastly, although our dataset has been verified as easily identifiable for humans by evaluators that span a balanced distribution across genders, we are interested in whether VLMs have any underlying gender biases that may affect their performance. 
Therefore, we only change the character to a female character and compare results. 
These results are shown in \autoref{tab:gender_eval}. 
As is the case in the angle variations, we also observe a lack of robustness in VLMs from the larger performance differences in the VLMs compared to that of humans. 
On a positive note, we do not observe a consistent preference for a particular gender by the VLMs.

\begin{table}[t]
    \centering
    \begin{adjustbox}{width=\linewidth}
\begin{tabular}{llcccccc}
\toprule
\multirow{2}{*}{Model} & \multirow{2}{*}{Method} & \multicolumn{3}{c}{MC} & \multicolumn{3}{c}{FF} \\
\cmidrule(lr){3-5} \cmidrule(lr){6-8}
 & & \male & \female & $\Delta \downarrow$ & \male & \female & $\Delta \downarrow$ \\
\midrule
\multirow{2}{*}{\makecell[l]{Qwen2.5 VL \\ (3B)}} & Zero-shot & 34.9 & 29.1 & \textcolor{darkgreen}{5.8} & 2.3 & 1.2 & \textcolor{darkgreen}{1.1} \\
 & CoT & 43.0 & 37.2 & \textcolor{darkgreen}{5.8} & 0.0 & 2.3 & \textcolor{darkred}{2.3} \\
\multirow{2}{*}{\makecell[l]{Qwen2.5 VL \\ (7B)}} & Zero-shot & 39.5 & 41.9 & \textcolor{darkred}{2.4} & 5.8 & 9.3 & \textcolor{darkred}{3.5} \\
 & CoT & 41.9 & 46.5 & \textcolor{darkred}{4.6} & 8.1 & 10.5 & \textcolor{darkred}{2.4} \\
\multirow{2}{*}{\makecell[l]{Phi-3.5}} & Zero-shot & 29.1 & 34.9 & \textcolor{darkred}{5.8} & 2.3 & 2.3 & \textcolor{darkgreen}{0.0} \\
 & CoT & 41.9 & 33.7 & \textcolor{darkgreen}{8.2} & 4.7 & 2.3 & \textcolor{darkgreen}{2.4} \\
\multirow{2}{*}{\makecell[l]{InternVL2.5 \\ (8B)}} & Zero-shot & 31.4 & 33.7 & \textcolor{darkred}{2.3} & 2.3 & 5.8 & \textcolor{darkred}{3.5} \\
 & CoT & 25.6 & 24.4 & \textcolor{darkgreen}{1.2} & 1.2 & 5.8 & \textcolor{darkred}{4.6} \\
 \arrayrulecolor[gray]{0.8} \midrule
\multirow{3}{*}{\makecell[l]{GPT-4o Mini}} & Zero-shot & 41.9 & 44.2 & \textcolor{darkred}{2.3} & 11.6 & 12.8 & \textcolor{darkred}{1.2} \\
 & CoT & 43.0 & 53.5 & \textcolor{darkred}{10.5} & 16.3 & 10.5 & \textcolor{darkgreen}{5.8} \\
 & Few-shot & 74.4 & 65.1 & \textcolor{darkgreen}{9.3} & 9.3 & 10.5 & \textcolor{darkred}{1.2} \\
\multirow{3}{*}{\makecell[l]{Gemini 1.5 \\Flash}} & Zero-shot & 52.3 & 47.7 & \textcolor{darkgreen}{4.6} & 19.8 & 20.9 & \textcolor{darkred}{1.1} \\
 & CoT & 54.6 & 52.3 & \textcolor{darkgreen}{2.3} & 22.1 & 19.8 & \textcolor{darkgreen}{2.3} \\
 & Few-shot & 57.0 & 59.3 & \textcolor{darkred}{2.3} & 13.9 & 22.1 & \textcolor{darkred}{8.2} \\
\arrayrulecolor{black} \midrule
Human & - & 99.6 & 98.5 & \textcolor{darkgreen}{1.1} & 89.5 & 90.3 & \textcolor{darkred}{0.8} \\
\bottomrule
\end{tabular}
    \end{adjustbox}
    \caption{Performance comparison on \ours for gender variations. $\Delta$ is shown in \textcolor{darkgreen}{blue} if \male  $-$ \female $\geq0$ and in \textcolor{darkred}{orange} otherwise. Similar to angle variation results, results for VLMs vary largely depending on the gender without a consistent performance advantage of a certain gender, while human performance is consistent.}
    \vspace{-0.5em}
    \label{tab:gender_eval}
\end{table}

\begin{table*}[h]
    \centering
    \adjustbox{max width=\textwidth}{
    \begin{tabular}{lr|cc|cccccccccc}
        \toprule
        \multirow{2}{*}{\textbf{Model}} & \multirow{2}{*}{\textbf{Method}} & \multicolumn{2}{c}{\textbf{\texttt{Base \& blank}}} & \multicolumn{2}{c}{\textbf{\texttt{Base \& $=$back.}}} & \multicolumn{2}{c}{\textbf{\texttt{Base \& $\neq$back.}}} & \multicolumn{2}{c}{\textbf{\texttt{\adv \& blank}}} & \multicolumn{2}{c}{\textbf{\texttt{\adv \& $=$back.}}} & \multicolumn{2}{c}{\textbf{\texttt{\adv \& $\neq$back.}}} \\
        \cmidrule(lr){3-4} \cmidrule(lr){5-6} \cmidrule(lr){7-8} \cmidrule(lr){9-10} \cmidrule(lr){11-12} \cmidrule(lr){13-14}
        & & \textbf{MC} & \textbf{FF} & \textbf{MC} & \textbf{FF} & \textbf{MC} & \textbf{FF} & \textbf{MC} & \textbf{FF} & \textbf{MC} & \textbf{FF} & \textbf{MC} & \textbf{FF} \\
        \midrule
\multirow{3}{*}{\makecell[l]{Qwen 2.5 VL\\ (3B)}}  
  & Zero-shot & 34.9 & 2.3 & 61.6 & 30.2 & 27.9 & 0.0 & 30.2 & 1.2 & 60.5 & 29.1 & 24.4 & 0.0 \\
  & CoT & \textcolor{darkgreen}{43.0} & \textcolor{darkred}{0.0} & \textcolor{darkred}{57.0} & \textcolor{darkred}{25.6} & \textcolor{darkgreen}{27.9} & \textcolor{darkred}{0.0} & \textcolor{darkred}{29.1} & \textcolor{darkred}{0.0} & \textcolor{darkred}{58.1} & \textcolor{darkred}{22.1} & \textcolor{darkgreen}{25.6} & \textcolor{darkred}{0.0} \\
  & FT$^{\dagger}$ & \textcolor{darkred}{31.6} & \textcolor{darkred}{0.0} & - & - & - & - & \textcolor{darkred}{22.0} & \textcolor{darkred}{0.0} & - & - & - & - \\  \arrayrulecolor[gray]{0.8} \midrule

\multirow{3}{*}{\makecell[l]{Qwen 2.5 VL \\ (7B)}} 
  & Zero-shot & 39.5 & 5.8 & 68.6 & 38.4 & 32.6 & 1.2 & 34.9 & 0.0 & 64.0 & 30.2 & 30.2 & 0.0 \\
  & CoT & \textcolor{darkgreen}{41.9} & \textcolor{darkgreen}{8.1} & \textcolor{darkred}{62.8} & \textcolor{darkred}{37.2} & \textcolor{darkred}{31.4} & \textcolor{darkgreen}{3.5} & \textcolor{darkred}{27.9} & \textcolor{darkred}{0.0} & \textcolor{darkred}{61.6} & \textcolor{darkred}{17.4} & \textcolor{darkred}{26.7} & \textcolor{darkgreen}{1.2} \\
  & FT$^{\dagger}$ & \textcolor{darkred}{36.8} & \textcolor{darkred}{0.0} & - & - & - & - & \textcolor{darkred}{25.0} & \textcolor{darkred}{0.0} & - & - & - & - \\ \midrule

\multirow{3}{*}{\makecell[l]{Phi 3.5 \\ (4.2B)}} 
  & Zero-shot & 29.1 & 2.3 & 73.3 & 27.9 & 31.4 & 8.1 & 44.2 & 0.0 & 72.1 & 27.9 & 36.1 & 5.8 \\
  & CoT & \textcolor{darkgreen}{41.9} & \textcolor{darkgreen}{4.7} & \textcolor{darkred}{64.0} & \textcolor{darkgreen}{30.2} & \textcolor{darkred}{24.4} & \textcolor{darkred}{1.2} & \textcolor{darkred}{31.4} & \textcolor{darkgreen}{1.2} & \textcolor{darkred}{59.3} & \textcolor{darkgreen}{30.2} & \textcolor{darkred}{33.7} & \textcolor{darkred}{2.3} \\
  & FT$^{\dagger}$ & \textcolor{darkred}{26.3} & \textcolor{darkred}{0.0} & - & - & - & - & \textcolor{darkred}{22.0} & \textcolor{darkred}{0.0} & - & - & - & - \\ \midrule

\multirow{2}{*}{\makecell[l]{InternVL2.5 \\(8B)}} 
  & Zero-shot & 31.4 & 2.3 & 57.0 & 26.7 & 22.1 & 2.3 & 25.6 & 2.3 & 59.3 & 20.9 & 30.2 & 2.3 \\
  & CoT & \textcolor{darkred}{25.6} & \textcolor{darkred}{1.2} & \textcolor{darkgreen}{60.5} & \textcolor{darkred}{23.3} & \textcolor{darkgreen}{32.6} & \textcolor{darkred}{2.3} & \textcolor{darkgreen}{26.7} & \textcolor{darkred}{1.2} & \textcolor{darkred}{52.3} & \textcolor{darkred}{15.1} & \textcolor{darkred}{23.3} & \textcolor{darkred}{0.0} \\ \arrayrulecolor{black} \midrule

\multirow{3}{*}{GPT-4o Mini} 
  & Zero-shot & 41.9 & 11.6 & 66.3 & 39.5 & 37.2 & 3.5 & 33.7 & 8.1 & 67.4 & 33.7 & 36.1 & 2.3 \\
  & CoT & \textcolor{darkgreen}{43.0} & \textcolor{darkgreen}{16.3} & \textcolor{darkgreen}{73.3} & \textcolor{darkgreen}{47.7} & \textcolor{darkgreen}{44.2} & \textcolor{darkgreen}{8.1} & \textcolor{darkgreen}{44.2} & \textcolor{darkred}{4.7} & \textcolor{darkred}{65.1} & \textcolor{darkgreen}{38.4} & \textcolor{darkred}{36.1} & \textcolor{darkred}{1.2} \\
  & Few-shot$^\dagger$ & \textcolor{darkgreen}{74.4} & \textcolor{darkred}{9.3} & \textcolor{darkgreen}{94.2} & \textcolor{darkred}{39.5} & \textcolor{darkgreen}{52.3} & \textcolor{darkred}{0.0} & \textcolor{darkgreen}{70.9} & \textcolor{darkred}{2.3} & \textcolor{darkgreen}{89.5} & \textcolor{darkgreen}{40.7} & \textcolor{darkgreen}{59.3} & \textcolor{darkred}{0.0} \\ \arrayrulecolor[gray]{0.8} \midrule

\multirow{3}{*}{\makecell[l]{Gemini 1.5 \\Flash}} 
  & Zero-shot & 52.3 & 19.8 & 68.6 & 51.2 & 37.2 & 12.8 & 44.2 & 8.1 & 75.6 & 46.5 & 36.1 & 3.5 \\
  & CoT & \textcolor{darkgreen}{54.7} & \textcolor{darkgreen}{22.1} & \textcolor{darkgreen}{69.8} & \textcolor{darkred}{48.8} & \textcolor{darkgreen}{40.7} & \textcolor{darkred}{11.6} & \textcolor{darkgreen}{48.8} & \textcolor{darkgreen}{9.3} & \textcolor{darkred}{74.4} & \textcolor{darkgreen}{51.2} & \textcolor{darkgreen}{41.9} & \textcolor{darkgreen}{7.0} \\
  & Few-shot$^\dagger$ & \textcolor{darkgreen}{57.0} & \textcolor{darkred}{14.0} & \textcolor{darkgreen}{72.1} & \textcolor{darkred}{41.9} & \textcolor{darkgreen}{46.5} & \textcolor{darkred}{10.5} & \textcolor{darkgreen}{48.8} & \textcolor{darkred}{4.7} & \textcolor{darkgreen}{77.9} & \textcolor{darkred}{39.5} & \textcolor{darkgreen}{44.2} & \textcolor{darkred}{0.0} \\ \arrayrulecolor{black} \midrule 

Human & - & 99.6 & 89.5 & 98.5 & 89.2 & 99.2 & 93.4 & 98.5 & 93.8 & 99.2 & 94.1 & 99.2 & 95.0 \\
        \bottomrule
    \end{tabular}
    }
    \caption{Results for various methods to improve performance on \ours. The table follows the same format as \autoref{tab:main-perturbations}. 
    $^\dagger$Refer to \textsection \ref{subsec:methods} for details on the experimental setup for few-shot and fine-tuning results for which the number of evaluation samples is smaller. 
    Accuracy is shown in \textcolor{darkgreen}{blue} for methods that is higher than the corresponding zero-shot score and in \textcolor{darkred}{orange} otherwise.  
    }
    \vspace{-1em}
    \label{tab:main-improvements}
\end{table*}

\section{Improving on \ours}
\label{sec:improving}

Given the poor performance of VLMs on \ours, we are interested in whether simple methods can surface VLMs' potential to understand mimed actions. 
In this section, we discuss our attempts to improve their performance via such methods.

\subsection{Methods}
\label{subsec:methods}

The methods that we explore are the following: 
(\textit{i}) \textbf{Chain-of-Thought} (CoT) is a method of producing a reasoning chain before making a final judgement. We ask the model to describe what it sees in detail and then provide its prediction~\cite{wei2022chain}. 
(\textit{ii}) \textbf{Few-shot in-context learning} (Few-shot): For models that support few-shot in-context learning, we select three samples from the base configuration of \ours with minimal overlap in enacted actions (shooting a soccer ball, fishing, playing violin) and provide them as in-context examples that the models can leverage to improve their predictions on the remaining samples.  
(\textit{iv}) \textbf{Fine-tuning}: Lastly, we experiment with fine-tuning to see if fine-tuning on a small amount of data containing mimed actions can help models generalize to unseen ones. Since \ours only contains 86 samples in total per configuration, we fine-tune (FT) our model using a 5-fold validation approach with a 36/14/36 train/validation/test split.
The details of these splits are present in Appendix \ref{appdx:finetuning_details}.
Fine-tuning is conducted separately for each task type (free-form, and multiple choice). 
Due to limited compute, we limit our fine-tuning experiments to the base configuration and \adv \texttt{+ blank} background configuration (\texttt{(a)} and \texttt{(h)} in \autoref{fig:mime-dataset-overview}) and the Qwen 2.5 VL (3B, 7B) models and Phi 3.5 (4.2B). 
Refer to Appendix \ref{appdx:prompt_details} for the few-shot and CoT prompts and Appendix \ref{appdx:finetuning_details} for further details of our fine-tuning setup.

\subsection{Improvement Results}

The main results of these preliminary methods are shown in \autoref{tab:main-improvements}. 
We observe that, apart from the API-based black box models, most methods do not lead to consistent and significant improvements over the results from zero-shot. 
One noticeable improvement is that of GPT-4o Mini when it is given few-shot examples, where results on most variations are boosted to over 50\% for MC.
While a smaller boost, we see a similar trend for Gemini 1.5 Flash. 
However, the performance for most cases still remain very low for FF, indicating that they continue to struggle without contextual information. 
Overall, our results demonstrate that there is ample room for improvement for VLMs to acquire an understanding of human gestures that is as robust as those of humans. 

\subsection{Failure Mode Analysis}
\label{ssec:failure_mode_analysis}
In order to understand where improvement opportunities lie, we analyze the modes of failure by Gemini 1.5 Flash with CoT on FF to examine the reasoning they generate for making predictions without contextual hints provided by multiple choice options. 
The reasoning serves as a proxy of what the VLM observes and thus analyzing it can surface the point of failure that needs to be corrected.
We want to know whether the models fail to correctly describe the shown action or whether they can accurately describe it but cannot interpret it as the intended mimed action, and also how much they are affected by the aligned and misaligned backgrounds. 
We manually categorize the modes of failure for predictions of the first three columns in \autoref{tab:main-improvements} (i.e., using the base character with blank, aligned, and misaligned backgrounds). 

We find that, with the blank background, Gemini 1.5 Flash generates a description of the shown mimed action that is only partially correct 54\% of the time and completely incorrect 16\% of the time (e.g., \textit{They wind up their arm as if holding a ball, then perform a throwing motion with their arm and hand extending forward} for arm curls). 
In 13\% of instances, the description is correct, but it is interpreted incorrectly, leading to an incorrect prediction (e.g., predicts bowling a ball after generating \textit{They start with a wind-up motion, bringing their arm back, then swing forward as if releasing a ball} for baseball pitch).
When shown an aligned background, 43\% of predictions that were incorrect with the blank background become correct predictions. 
With the misaligned background, 24\% of all predictions are confused by irrelevant context provided by the background (e.g., predicts conducting an orchestra for climbing given a concert hall background), which leads to a drop in proportion of predictions that had completely or partially correct descriptions of the shown mimed action.\footnote{
We share the full statistics and more examples of each mode of failure in Appendix ~\ref{appdx:experimental_result_details}.
Other VLMs also show similar failure patterns, with most instances of failure caused by partial or completely incorrect descriptions of the mimed actions. 
Overall, these results indicate that future research should prioritize training VLMs that can accurately describe the human gestures they observe. 
}

\section{Related Work}
\label{sec:related_work}

\subsection{Nonverbal Communication}

One major branch of NVC research leverages NVC to enhance predictions for a downstream task, such as using posture, prosodic features, and facial expressions to predict dialogue acts~\cite{Sridhar2009CombiningLS, boyer-etal-2011-affect, ha-etal-2012-combining}, gaze, head movement, and breath patterns to detect turn-taking and engagement behavior~\cite{jokinen-2010-non, Ishii2013PredictingNS, Ishii2014AnalysisAM, Ishii2015PredictingNS, Ishii2016PredictionOW, Ishii2016UsingRT}, visual information and motion capture data for emotion representations and predictions~\cite{busso2008iemocap, zhang2023learning}.
There are a few prior work that seeks to predict NVC with gestures, but they are constrained to those expressed with limited body parts, such as hands for hand gestures  
~\cite{10.1109/TRO.2015.2475956, Kapitanov_2024_WACV} or sign language~\cite{Papastratis2021ContinuousSL, kezar-etal-2023-improving}.  
Others use verbal signals to predict or generate NVC, usually in the context of developing realistic virtual agents~\cite{Graf2002VisualPF, Busso2007RigidHM} or robots~\cite{shamsuddin2011humanoid, Sakai2015OnlineSH, Cass2018AFT}, such as using dialogue acts and affective information to predict nods ~\cite{ Lee2010PredictingSH,  ishii-etal-2018-predicting}.  
In contrast, \ours examines whether machine learning models have a robust understanding of explicit full-body gestures, a fundamental prerequisite to comprehending more variable and subtle gestures in the full spectrum of NVC, by evaluating whether they can identify mimed actions --- a subset of NVC with low interpretation variability.

\subsection{Action Recognition} 

While mimed action understanding is an instrumental step towards general NVC understanding, it is also highly related to action understanding. 
Many datasets exist for evaluating whether machine learning models understand human actions~\cite{kong2022human, sun2022human}, with early work focused on sporting actions~\cite{kuehne2011hmdb, soomro2012ucf101, karpathy2014large, idrees2017thumos} and recent work expanding to a larger scope and scale, including daily activities that are crowdsourced~\cite{sigurdsson2016hollywood, damen2018scaling} and extracted from YouTube~\cite{Heilbron2015ActivityNetAL, xu2016msr, krishna2017dense, kay2017kinetics, sanabria2018how2, zhou2018towards,  miech2019howto100m, weinzaepfel2021mimetics}, Tumblr~\cite{li2016tgif}, Flickr~\cite{anne2017localizing}, or movies~\cite{torabi2015using, rohrbach2015dataset, rohrbach2017movie}.  
~\citet{Miech2020RareActAV} takes a step further to evaluate action recognition robustness with a dataset that contains rare activities (e.g., blending phone and cutting keyboard).   
While these datasets collectively cover more than hundreds of different classes, none of them contain mimed actions. 
As such, even if VLMs perform well on these datasets, it is unclear whether they have a robust understanding of complex human body motions or if they are relying on spurious correlations provided by the salient context. 

The CMU-MMAC Database~\cite{Torre2008GuideTT} is similar to \ours in that it contains animated videos of motion capture data, but the animations are extremely simple, featuring a black background with yellow stick figures, and do not include finger and thumb joints, which lowers fidelity to the captured actions.
IEMOCAP~\cite{busso2008iemocap} is also based on motion capture data but it is focused on emotion prediction in dyadic conversations 
and motion capture is only collected for the face, head, and hands.  
\citet{van2017production}~presents a dataset of footage of human participants pantomiming various objects, but it is limited to hand gestures. 
Lastly, the Mimetics dataset~\cite{weinzaepfel2021mimetics} is the most similar to \ours in that in contains live action footage videos of mimed actions extracted from YouTube with varying amounts of relevant context provided. 
However, it lacks the flexibility to systematically adjust the amount of relevant context provided in each video with precision for conducting the ablative analysis possible with \ours. 
Moreover, the Mimetics dataset may be included in the training of VLMs as the videos are from YouTube, and therefore \ours serves as a more reliable benchmark for mimed action understanding unaffected by data leakage.

\section{Conclusion}
\label{sec:conclusion}
We introduce \ours, a novel video-based question answering benchmark that consist of animations of 86 mimed actions created with motion capture data and 3D graphics software. 
\ours contains systematic perturbations in character, background, and viewpoint to assess the robustness of VLMs' understanding of full-body gestures. 
While humans demonstrate almost perfect accuracy and remain highly robust to all modifications in \ours, both open-weight and API-based VLMs struggle, particularly in the free-form format where recognition accuracy approaches zero under adversarial perturbations. 
These results highlight the need for further research in enhancing VLMs with a robust understanding of human gestures to establish a crucial bedrock for NVC comprehension.  

\section*{Acknowledgment}

We thank David Nelson and the Mixed Reality Lab at the Institute for Creative Technologies for providing access to their Vicon system, which was an indispensable equipment for collecting the motion capture data used in creating \ours.  

\section*{Limitations}
\label{sec:limitations}

\paragraph{Lack of Photorealism}
The main limitation of \ours is that it is not photorealistic as it contains animated videos of motion capture data. 
This lack of realism may introduce a domain shift for VLMs for which the majority of the training data is likely to be live action footage rather than animations, leading to an artificially discounted performance of VLMs on \ours.
However, given that humans can successfully interpret mimed actions in \ours, we argue that models should also be able to achieve comparable performance to humans on \ours if they develop a robust understanding of human gestures by generalizing from what they learn through more photorealistic content. 

The flip side of this concern is that  performance on \ours do not translate to equivalent performance on mimed actions captured as live action footage. 
We believe that this concern is addressed to a reasonable extent in that \ours contains multiple perturbations of the same set of actions with varying characters, backgrounds, and viewpoints. It would be unlikely for a VLM to achieve high accuracy on all of these variants without a robust understanding of human actions that do not translate to understanding of these actions in live action footage. 
In addition, as the fidelity of digital assets improve and with the availability of more compute, we will be able to create versions of \ours that are increasingly photoreaslitic, which further mitigates this concern. 
Therefore, despite concerns arising from the lack of photorealism, we argue that \ours is the most advantageous for systematic analysis of robustness because of the ease of producing variants that enable ablation studies. 
Alternative methods, i.e., using live action footage or video generation models, are significantly limited in being able to modify equivalent mimed actions at the same level of flexibility and consistency. Refer to a detailed discussion on alternative methods in Appendix \ref{appdx:alternative_generation}.  

\paragraph{Representation Bias}
Next, \ours only consists of animations that are based on motion capture data from two actors. 
However, one of these actors is an Asian male non-professional actor and the other is an European female professional actor. 
Despite minimal overlap in demographics of these two actors, our human evaluation results show that there is not a higher recognition accuracy for samples that come from one actor over those of another.\footnote{99.0\% vs. 98.8\% in multiple choice format and 92.6\% vs. 90.6\% for free-form format. A t-test indicates insignificant differences in these accuracies at $p<0.01$ ($p=0.71$ and $p=0.03$, respectively).} 
In other words, there may be differences in how people enact actions as mimes, but they are not significant enough to affect recognition, at least for the pool of 60 participants that we recruited. 
We believe these results should address concerns of representation bias of the mimed actions in \ours. 

\paragraph{Data contamination in \realdataset}
Another caveat of our results that compare VLM performance between \ours and \realdataset is that samples in \realdataset may have been included in the training of the VLMs that we evaluated, thus inflating their results on \realdataset due to data contamination. 
Also, an ideal systematic study of a model's understanding of actions would have entailed studying \realdataset without contextual information provided by the background, but we could not pursue this path due to technical challenges preventing background removal in \realdataset. 
Unfortunately, the VLMs that we study in this paper do not share the full scope of their training data, and therefore we cannot confirm whether strong performance on \realdataset is due to data leakage or because they can reliably identify actual actions (as opposed to mimed actions) from live action footage. 

\paragraph{Fine-tuning result caveats}
A noteworthy limitation in our improvement results from \textsection \ref{sec:improving} is that our fine-tuning experiments do not provide conclusive evidence regarding the effectiveness of fine-tuning for improving model performance on \ours. 
Our fine-tuning experiments are an attempt at domain adaptation using a limited sample size, which likely leads to overfitting, preventing the model from achieving meaningful generalization. 
While this does not rule out the potential benefits of fine-tuning on larger and more diverse datasets, our findings suggest that additional research is necessary to explore optimal fine-tuning strategies for tasks as challenging as \ours.

\section*{Ethical Considerations}
\label{sec:ethical_considerations}

While \ours serves as an important milestone for VLMs to reach on their path to commanding fluent NVC, strong performance on \ours should be interpreted with caution. 
First, it is important to note that the set of mimed actions explored in \ours is not exhaustive of actions that can be possibly mimed. 
It is a carefully curated subset that we find high agreement among human participants with diverse backgrounds and therefore propose as one of the lowest-hanging fruits in NVC recognition. 
In other words, strong performance on \ours should be considered a prerequisite being met for VLMs that can be further improved to understand and also generate the more nuanced forms of NVC. 
It should not be interpreted incorrectly as an indication that VLMs can command NVC fluently and thus is ready to be applied to downstream tasks that require such skills. 
Future work should expand the scope on \ours to include more nuanced gestures that potentially have relatively lower universal agreement but nonetheless have high intracultural agreement.

In addition, with concerns of test data leakage on the rise ~\cite{zhou2023don, xu2024benchmarking, jiang2024investigating}, performance improvement on \ours may not be indicative of robust understanding of human gestures if improvement on \ours is achieved simply by training more on a data distribution similar to \ours instead of improving the generalizability of VLMs.
Therefore, it is important for VLM developers to be cautious and transparent about how they source visual data to prevent misleading performance gains on \ours. 
Luckily, our pipeline for creating \ours can be easily replicated for producing unseen permutations of mimed actions in \ours to further test the limits of generalizability if there is suspected data contamination.

\bibliography{anthology,custom}
\appendix

\section*{Appendix}

\section{\ours Details}
\label{appdx:dataset_details}

\subsection{Motion Capture Technical Details}
\label{appdx:motion_capture_details}

We collect motion capture with actors wearing motion capture suits configured in the Vicon 10 finger marker setup, in addition to the standard 53 body marker setup.\footnote{\url{https://help.vicon.com/space/Shogun112/31229851/Place+markers+on+a+performer}} 
Motion capture is performed on a Vicon stage configured with Vero capture cameras driven by Vicon Shogun 1.11. 
An example of a single frame from the resulting motion capture data is shown in (1) of \autoref{fig:data-collection-overview}. 
Finally, the dataset is batch cleaned, post-processed, and exported via Shogun Post into FBX format for further processing in Blender.

\subsection{Blender Macro Script}
\label{appdx:blender-processing}

Our script imports the character and motion capture armature, adjusting their resting positions to be as aligned as possible using the MCATS plugin,\footnote{\url{https://github.com/absolute-quantum/cats-blender-plugin}} and using the Rokoko Studio Live plugin\footnote{\url{https://github.com/Rokoko/rokoko-studio-live-blender}} to retarget the animations from the motion capture data to the character.
In addition, a sun light source and large plane at the feet level of the character are added for shadow capture for more realistic videos. 
Lastly, a camera is added so that videos can be rendered from the camera's viewpoint. 
We select a conservatively zoomed out viewpoint in order to make sure that the full action sequence is captured in the rendered output.  

\subsection{Render settings}

Each frame is rendered with the following rendering configurations:
\begin{itemize}
\setlength{\itemsep}{0pt}
\setlength{\parskip}{0pt}
\setlength{\parsep}{0pt}
    \item Number of samples: 32 
    \item Maximum number of light bounces: 1
    \item Resolution: 1280 $\times$ 720
    \item Adaptive threshold: 0.5
    \item Denoise using GPU: True 
    \item Use persistent data: True
    \item Caustics reflective: False
    \item Caustics refractive: False 
    \item Use light tree: Falses
\end{itemize}
We find these settings to strike a reasonable balance between video quality and render time. 

We process rendering jobs in parallel on P100 and V100 GPUs, depending on availability. 
The final step of overlaying the frames with transparent backgrounds over various backgrounds are accelerated with \texttt{parallel}\cite{tange_2024_11247979}. 
Generative AI workloads were run locally on an RTX 4090.

\section{Prompt Details}
\label{appdx:prompt_details}

We provide the templates for our prompts here: 
\subsection{Zero-shot Multiple Choice}
{\small
\begin{verbatim}
What action is the person miming in this 
image/video?
Choose the most accurate description from the 
options below.

A. {options[0]}
B. {options[1]}
C. {options[2]}
D. {options[3]}

Respond with just a single letter (A, B, C, or D).
\end{verbatim} 
}

\subsection{Zero-shot Free-form}
{\small
\begin{verbatim}
What action is the person miming in this 
image/video?
Describe the action in a single short phrase 
(under 5 words). 

You can think out the action in a chain of thought, 
but please reply on the final line of your 
response, a single short phrase (under 5 words).

This action is being 'mimed' meaning backgrounds 
or objects that are relevant may not be present. 
Think about only the *action* taking place 
in the video, and give a response for what it looks
like the character is "acting out" or 
doing "charades" of.
\end{verbatim} 
}

\subsection{CoT Multiple Choice}
{\small
\begin{verbatim}
What action is the person miming in this 
image/video? 
Choose the most accurate description from the
options below.

A. {options[0]}
B. {options[1]}
C. {options[2]}
D. {options[3]}


Carefully think through the answer, by detailing
the particular actions and movements that you see 
the person doing. Your output should contain your
explanation, and then on a new line, a single 
letter corresponding to the answer you choose, with 
no punctuation. An example response is shown below:

'In the video, the person is moving a single 
arm back and forth, as if they are swinging a bat. 
This action is most accurately described by 
option B.

B'
\end{verbatim} 
}

\subsection{CoT Free-form}
{\small
\begin{verbatim}
What action is the person miming in this 
image/video? 
Carefully think through the answer, by detailing
the particular actions and movements 
that you see the person doing. 

This action is being 'mimed' meaning backgrounds
or objects that are relevant may not be present.
Think about only the *action* taking place
in the video, and give a response for what it
looks like the character is "acting out" or
doing "charades" of. Your output should contain
your explanation, and then on a new line,
a short phrase (under 5 words) corresponding to
your answer, with no punctuation or answer 
prefix such as 'Answer:'
\end{verbatim} 
}

\subsection{Few-shot ICL Multiple Choice}
{\small
\begin{verbatim}
What action is the person miming in this video? 
Choose from:
A. {options[0]}
B. {options[1]}
C. {options[2]}
D. {options[3]}
Answer with just a single letter (A, B, C, or D).
Answer: <answer>

What action is the person miming in this video? 
Choose from:
A. {options[0]}
B. {options[1]}
C. {options[2]}
D. {options[3]}
Answer with just a single letter (A, B, C, or D).
Answer: <answer>
...
What action is the person miming in this video? 
Choose from:
A. {options[0]}
B. {options[1]}
C. {options[2]}
D. {options[3]}
Answer with just a single letter (A, B, C, or D).
\end{verbatim} 
}

\subsection{Few-shot ICL Free-form}
{\small
\begin{verbatim}
What action is the person miming in this video? 
Describe the action in a single short phrase.
Answer: <answer>
What action is the person miming in this video? 
Describe the action in a single short phrase.
Answer: <answer>
...
What action is the person miming in this video?
Describe the action in a single short phrase.
\end{verbatim} 
}

\section{Fine-tuning Details}
\label{appdx:finetuning_details}

The $n=5$ folds that we use for N-fold training for fine-tuning experiments are shown in \autoref{tab:finetuning-folds}.
During FT, only the vision encoder is trained, while the text encoder remains frozen. We train for 7 epochs with an initial learning rate of 2$e$-5, following a cosine learning rate schedule. 
The batch size is set to 8, and we use the AdamW optimizer with $\beta_1 = 0.9$ and $\beta_2 = 0.999$. 
To optimize the balance between computational speed and precision, BF16 and TF32 are enabled. All models are trained using 2× A100 GPUs.

\begin{table*}
    \centering
    \begin{adjustbox}{width=\linewidth}
    \begin{tabular}{ccccc}
        \toprule
        Fold 1 & Fold 2 & Fold 3 & Fold 4 & Fold 5 \\
        \midrule
  \texttt{Volleyball001} & \texttt{Climbing001} & \texttt{DrinkingCoffee001} & \texttt{ConsoleGaming01} & \texttt{ArmCurls001} \\
        \texttt{VolleyballServe} & \texttt{Climbing01} & \texttt{ShootingAHandgun001} & \texttt{Darts001} & \texttt{ArmCurls01} \\
        \texttt{WeightedSquat002} & \texttt{DeadLift001} & \texttt{ShootingARifle001} & \texttt{Bowling003} & \texttt{ArmCurls03} \\
        \texttt{CheckingWatch001} & \texttt{Deadlift01} & \texttt{ShootingHandgun01} & \texttt{Bowling01} & \texttt{Baseball004} \\
        \texttt{CheckingWatch01} & \texttt{Archery001} & \texttt{Basketball001} & \texttt{Weightlifting001} & \texttt{BaseballPitch002} \\
        \texttt{Swimming001} & \texttt{Archery01} & \texttt{BasketballLayup001} & \texttt{Violin002} & \texttt{BaseballPitch02} \\
        \texttt{Swimming002} & \texttt{Driving002} & \texttt{BasketballLayup02} & \texttt{ShotPut001} & \texttt{CheckingPhone002} \\
        \texttt{Swimming03} & \texttt{Driving003} & \texttt{BasketballShot02} & \texttt{ShotPut01} & \texttt{WatchingTV01} \\
        \texttt{Swimming04} & \texttt{Soccer003} & \texttt{Boxing001} & \texttt{DrivingSitting001} & \texttt{SittingAndWriting001} \\
        \texttt{Swimming06} & \texttt{SoccerShot01} & \texttt{Boxing03} & \texttt{DrivingSittingDown03} & \texttt{TakingPhotoWithCamera001} \\
        \bottomrule
    \end{tabular}
    \end{adjustbox}
        \caption{
    The action IDs in \ours that are divided into five folds we use for our fine-tuning setup. 
    }
    \label{tab:finetuning-folds}
\end{table*}

\section{Human Evaluation Details}
\label{appdx:human_evaluation_details}

\begin{figure*}
    \centering
    \includegraphics[width=\linewidth]{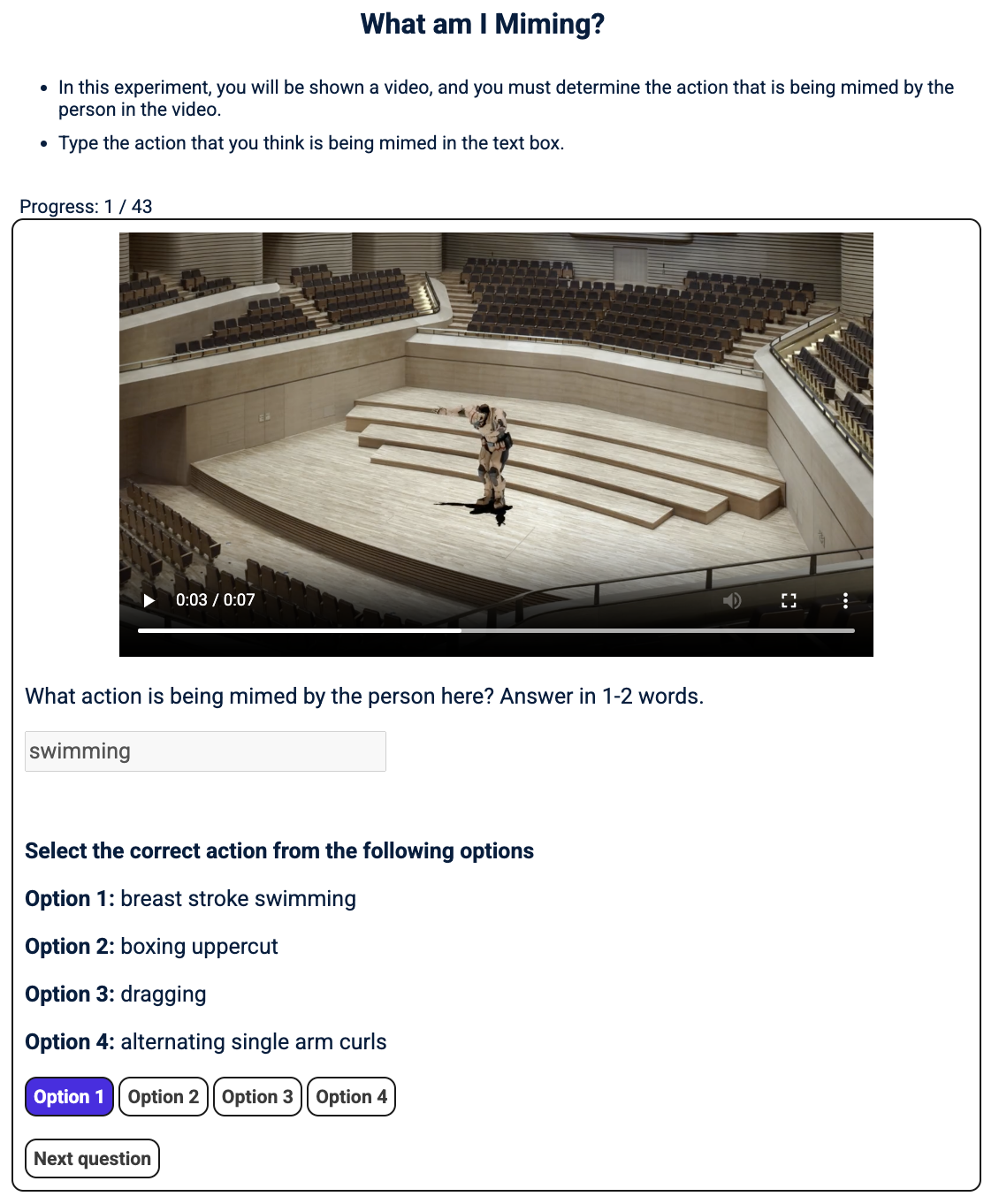}
    \caption{Our interface for human evaluation. The evaluators can only attempt to answer the question after seeing the full video. After answering a free-form short answer question, they are asked to complete a multiple choice equivalent before moving on to the next sample.}
    \label{fig:human_evaluation_interface}
\end{figure*}

Without any prior guidance, each participant is asked to answer the question in free-form format first after watching a sample in \ours and then answer the same question with the multiple choice format.
The interface shown to our participants is illustrated in \autoref{fig:human_evaluation_interface}.
While this setup is efficient for collecting both free-form and multiple choice format results from a single participant, options shown for the multiple choice format in prior samples may provide contextual information for free-form format answers in the remaining samples. 
However, we find this effect to be negligible in comparison to the large performance gap between humans and models: 
human performance with the free-form format on the first samples that they annotate is \char`\~88\% while that of the remaining samples is \char`\~93\%.   
Each participant annotates half of the total samples (43) in one of the configurations shown in \autoref{fig:mime-dataset-overview} in order to keep annotator burden low.

\section{Failure Mode Anlaysis Details}
\label{appdx:experimental_result_details}
\setlength{\fboxsep}{0pt}
\renewcommand{\arraystretch}{1.3} 

\begin{table*}[h]
    \centering
    \adjustbox{max width=\textwidth}{
    \begin{tabular}{ccrp{3cm}p{3cm}p{10cm}}
    \toprule 
        Configuration & Mode & \multicolumn{1}{c}{\%} & Mimed Action & Prediction & Generated CoT Description \\ \midrule
        \multirow{1}{*}{\texttt{Base +blank}} & \Circle & 16.3 & Drinking out of a cup & Answering a phone call & \textcolor{darkred}{They bring their hand to their ear as if holding a phone, and move their mouth as if talking.  They also have a slightly downward gaze, as if focusing on the phone.} \\ 
        & \RIGHTcircle & 53.5 & Driving & Riding a tiny board & \textcolor{darkred}{They are miming the action of holding a small object, such as a remote control,} and are \textcolor{darkgreen}{making small, precise movements with their hands as if they are controlling or manipulating something. Their body posture and stance remain relatively static.}  The actions are subtle, suggesting a delicate or precise control of the imagined object. \\
        & \CIRCLE  & 12.8 & Pushing & Falling down &  \textcolor{darkgreen}{They bend their knees, lean forward, and spread their arms out} as if they are losing their balance and falling onto the ground. Their posture and movements strongly suggest a loss of equilibrium resulting in a fall. The shadow on the ground further emphasizes this interpretation. \\ 
        & \checkmark & 3.5 & Backstroke swimming & Falling down & \textcolor{darkgreen}{The person in the video is lying on their back. Their arms are initially by their sides, then they slowly raise their arms above their head, then bring them back down to their sides.  Their legs are spread slightly apart. The movements suggest the person is acting out falling and lying on the ground.} \\  \midrule 
        \multirow{1}{*}{\texttt{Base =back.}} & \Circle & 18.6 & Baseball swing & Pitching a baseball & \textcolor{darkred}{They go through the windup motion, bringing their arm back and then forward in a throwing motion. Their legs and body also shift their weight from one foot to the other, mimicking the motion of a pitcher.} \\ 
        & \RIGHTcircle & 23.3 & Playing piano & Pulling something heavy & \textcolor{darkgreen}{The person in the video is crouched down, and moving their arms and hands} \textcolor{darkred}{in a way that mimics pulling something heavy.}  Their body language suggests exertion and strain.  \\
        & \CIRCLE  & 7.0 & Frontstroke swimming & Belly flopping & \textcolor{darkgreen}{The person in the video is lying on their stomach on the sand, their legs extended behind them, and their arms outstretched in front. Their body is relatively flat against the ground. One arm is slightly raised, possibly suggesting a wave-like motion. The overall posture and movement suggest a person mimicking the action of someone swimming, or more specifically, diving or belly-flopping into the water.} \\ 
        & \lightbulb & 37.2 & Throwing dart & Throwing darts & They go through the motions of holding an imaginary dart, aiming, and then releasing it with a follow-through. \\  \midrule 
        \multirow{1}{*}{\texttt{Base $\neq$back.}} & \Circle & 14.0 & Drinking out of a cup & Looking through binoculars & \textcolor{darkred}{They hold their hands up to their eyes in a characteristic manner, mimicking the way one would hold and adjust binoculars to look through them. Their hand movements are consistent with focusing and looking through an imaginary device.} \\ 
        & \RIGHTcircle & 38.4 & Throwing dart & Answering phone call & They \textcolor{darkred}{bring their hand to their ear as if holding a phone}, and \textcolor{darkgreen}{then extend their arm and point} as if speaking to someone on the other end of the call. \\
        & \CIRCLE  & 10.5 & Deadlifting & Picking up a spill & \textcolor{darkgreen}{The person in the video is bending over, appearing to pick up something from the floor.  Their hands move as if they are gathering or collecting a substance. Their body is positioned as if they are carefully handling whatever is on the ground.} \\ 
        & \confused & 24.4 & Climbing & Conducting orchestra & \textcolor{darkred}{They are standing in a relatively formal posture, and their arm movements are sweeping and expressive, suggesting they are leading a musical piece}.  \textcolor{darkgreen}{The upward and downward motions of their arms} \textcolor{darkred}{mimic the conducting of different musical sections or instruments. The movements are deliberate and rhythmic, further supporting the interpretation of conducting.} \\  
    \bottomrule 
    \end{tabular}
    }
    \caption{Gemini 1.5 Flash with CoT's prediction category distributions (*Note \% indicates proportion of predictions, including correct ones) and corresponding examples, as discussed in \textsection \ref{ssec:failure_mode_analysis}. \Circle: completely incorrect description. \RIGHTcircle: partially correct description of action. \CIRCLE: correct description with only minor discrepancies but incorrect interpretation. \checkmark: correct description and valid alternative interpretation. 
    Correct descriptions are in \textcolor{darkgreen}{blue} and incorrect descriptions are in \textcolor{darkred}{orange}.}
    \label{tab:cot_result_examples}
\end{table*}

Distribution of failure modes and examples of each mode are shown in \autoref{tab:cot_result_examples}.

\section{Alternative Methods for Creating \ours}
\label{appdx:alternative_generation}

\subsection{Live Action Footage}
\label{appdx:real_footage}

The ideal setup that does not introduce domain shift is to create an equivalent of \ours that contains live action footage. 
However, we intentionally avoid this option because of the difficulty to create systematic variations of the same mimed actions that enable robustness analysis and concerns of privacy of the actors that would be included in said dataset. 
Our attempts with removing objects and replacing backgrounds in each frame of the videos in \realdataset produced inconsistent results, and even if they were consistent, we would need another method to perturb the actor in a way that the resulting footage remains photorealistic. 

\subsection{Video Generation Models}
\label{appdx:video_generation_models}
\begin{figure*}[htbp]
    \centering
    \begin{subfigure}[t]{0.45\linewidth}
        \centering
        \includegraphics[width=\linewidth]{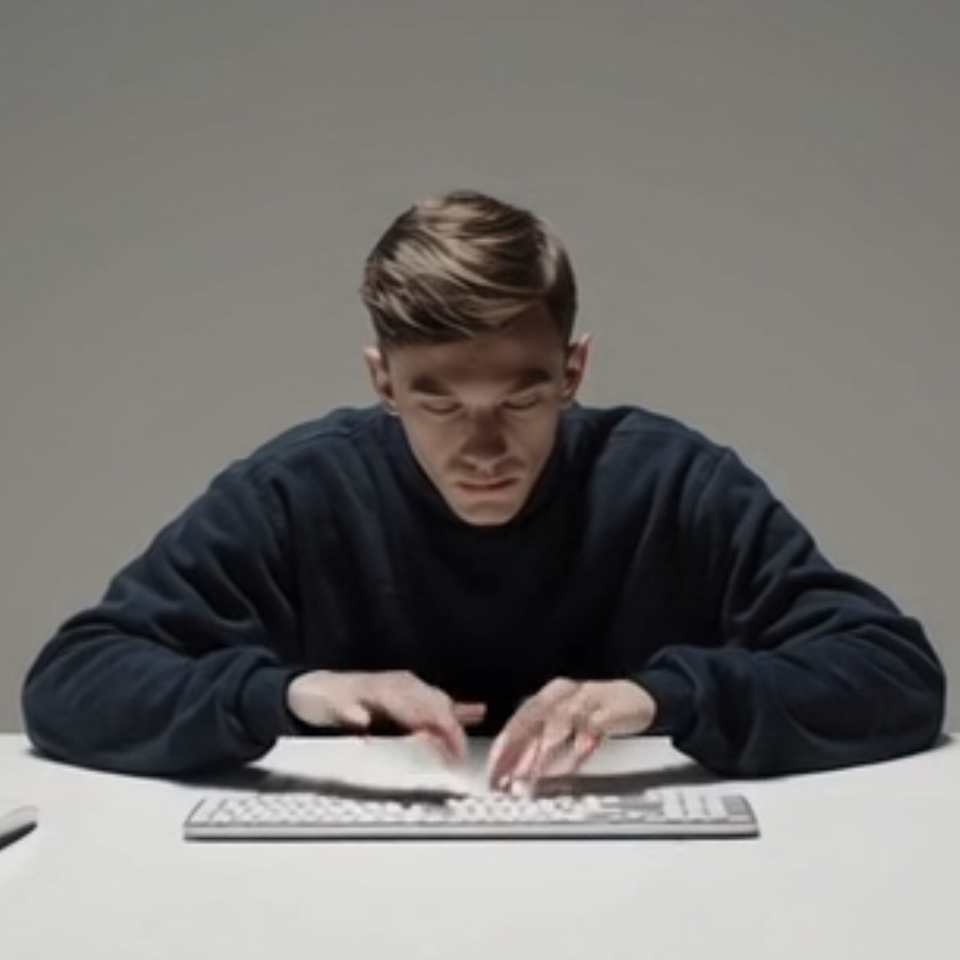}
        \caption{OpenAI Sora's output with prompt: \texttt{still shot without background of someone miming typing sitting by a desk without any objects on it}.}
        \label{fig:sora-1}
    \end{subfigure}
    \hfill
        \begin{subfigure}[t]{0.45\linewidth}
        \centering
        \includegraphics[width=\linewidth]{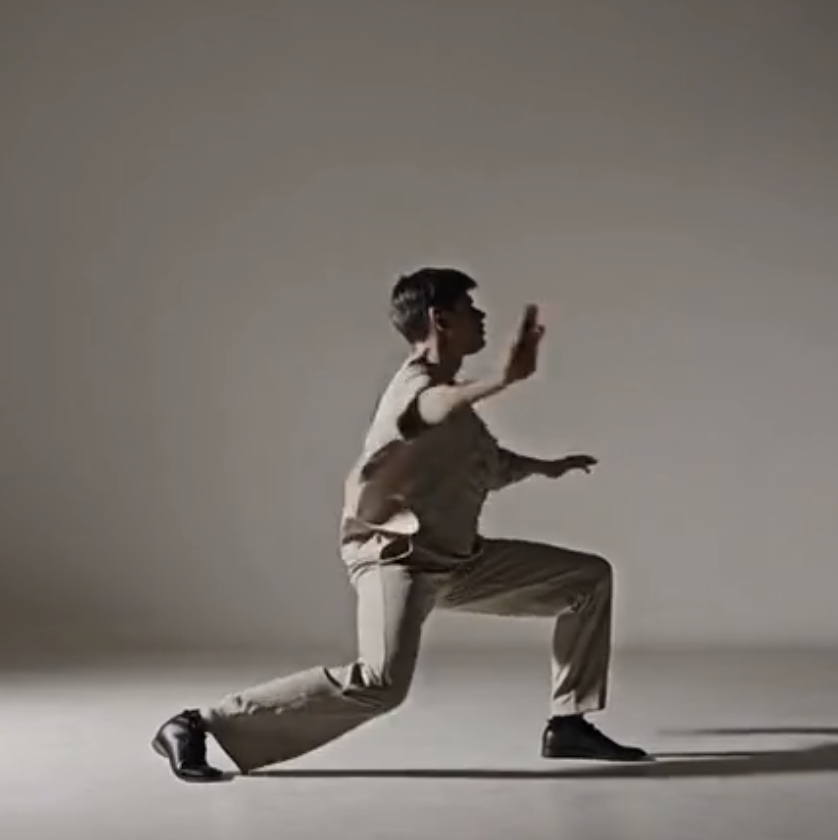}
        \caption{OpenAI Sora's output with LM-generated prompt: \texttt{Generate a high-quality video of a person performing mime movements that resemble fencing. The individual should use expressive body language, dynamic footwork, and precise hand gestures to create the illusion of fencing without any actual fencing equipment, such as swords or protective gear. The performance should be fluid and theatrical, emphasizing exaggerated parries, lunges, and ripostes to convey the essence of fencing through mime alone. The person should be dressed in neutral or casual clothing suitable for a performance, with a simple background that keeps the focus on their movement.}}
        \label{fig:sora-2}
    \end{subfigure}
    \hfill
    \begin{subfigure}[t]{0.45\linewidth}
        \centering
        \includegraphics[width=\linewidth]{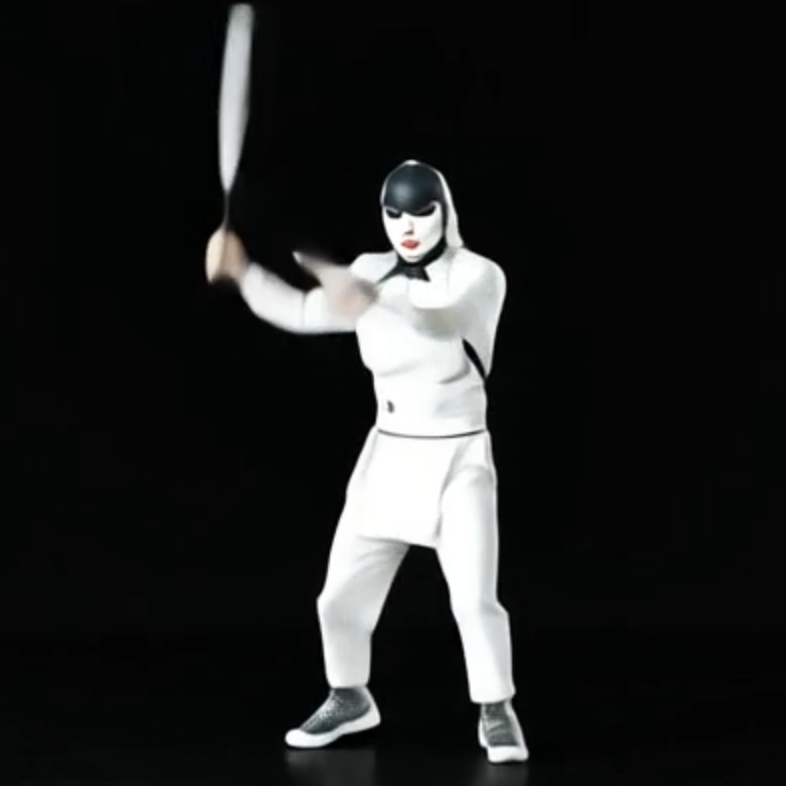}
        \caption{Runway's output with prompt: \texttt{Generate a video of a person miming a fencing match without any fencing equipment. The person should perform precise exaggerated fencing movements such as lunges, parries, and ripostes. Their footwork should be light and agile, moving back and forth as if engaged in a real bout. }}
        \label{fig:runway}
    \end{subfigure}
    \hfill
    \begin{subfigure}[t]{0.45\linewidth}
        \centering
        \includegraphics[width=\linewidth]{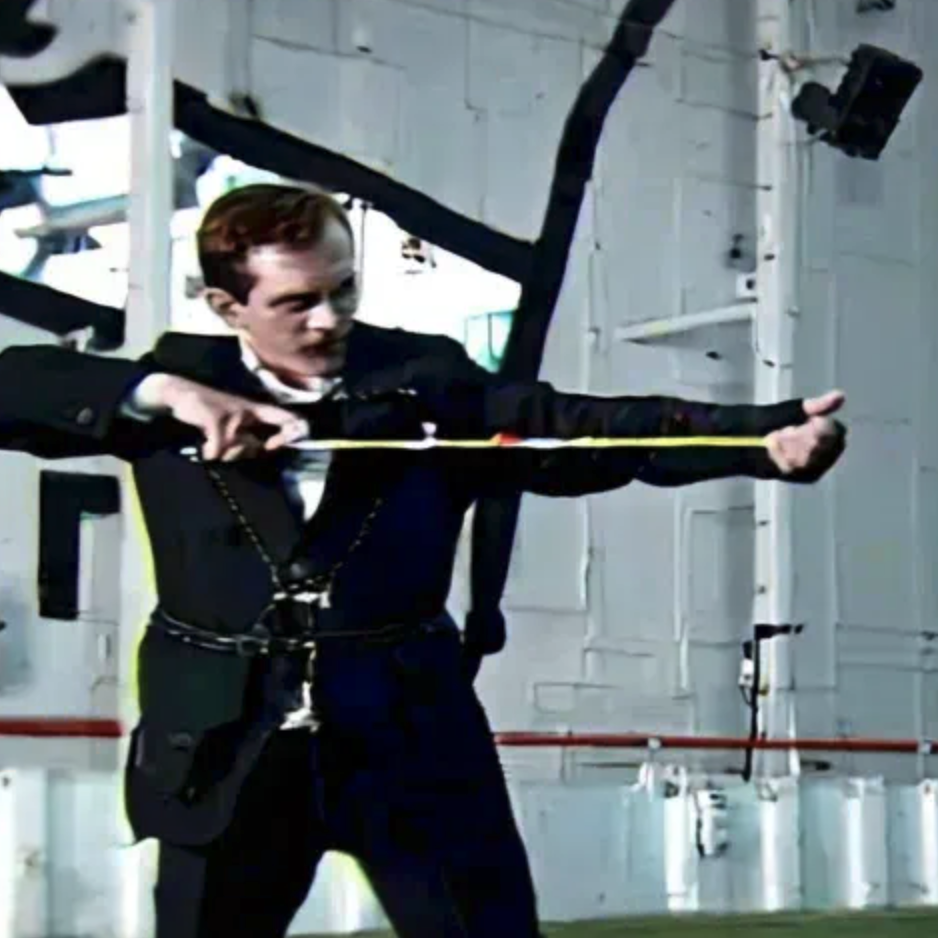}
        \caption{Hunyuan-Large's~\cite{sun2024hunyuanlargeopensourcemoemodel} output with prompt: \texttt{Man acting like shooting an arrow without anything in his hands. This should be a mimed action without any props.
}}
        \label{fig:hunyuan}
    \end{subfigure}
    \caption{Snapshots of outputs from various video generation models to generate mimed actions. All models that we tested failed to produce videos that either did \textbf{not} include the action's key object (e.g., keyboard while typing, bow and arrow while shooting an arrow) or correctly act out the intended action.}
    \label{fig:video-generation-model-outputs}
\end{figure*}

We also explore video generation models for creating \ours and show sample outputs in \autoref{fig:video-generation-model-outputs}. 
For paid services, we test Sora\footnote{\url{https://openai.com/sora/}} and Runway\footnote{\url{https://runwayml.com/}}, and for open-weight models, we use  a variety of Hunyuan \cite{sun2024hunyuanlargeopensourcemoemodel} fp16 and bf8 models using ComfyUI's\footnote{\url{https://github.com/comfyanonymous/ComfyUI}} recommended text-to-video Hunyuan workflow ~\cite{kong2024hunyuanvideo}. 
Despite various prompts, all video models struggle to generate mimed actions and generate the action with the salient context still present in the video, even when explicitly asked not to include it (see ~\autoref{fig:runway}) or ensuring it is not mentioned in the prompt (see ~\autoref{fig:sora-1}.
We also try prompts that are generated by language models, such as the output for the prompt: \texttt{``Generate a prompt for a video generation model to generate a video of someone miming fencing such that the resulting video  does not include any fencing equipment''}.
While this avoids producing salient context in some cases, it fails to produce a video that matches the intended action (e.g., dancing move shown for a prompt for fencing \autoref{fig:sora-2}).  

\end{document}